\documentclass[runningheads]{llncs}
\usepackage[T1]{fontenc}
\usepackage{comment}
\usepackage[table]{xcolor}
\usepackage{colortbl}
\usepackage{graphicx}
\usepackage{hyperref}
\usepackage{listings}
\usepackage{xcolor}
\usepackage{cite}
\usepackage{amsmath}
\usepackage{algorithm}
\usepackage{algpseudocode}
\usepackage{booktabs}
\usepackage{multirow}
\usepackage{afterpage}
\usepackage{amsmath}
\usepackage{algorithm}
\usepackage{algpseudocode}
\usepackage{amsfonts}
\usepackage{subcaption}

\newcommand{\fref}[1]{Fig.~\ref{#1}}

\usepackage[misc]{ifsym}
\usepackage{orcidlink}
\usepackage{graphicx,wrapfig,lipsum}
\usepackage{threeparttable}
\usepackage[margin=1.575in]{geometry}

\begin{document}
\title{Unlocking Neural Transparency: Jacobian Maps for Explainable AI in Alzheimer's Detection}
\titlerunning{Jacobian Maps for Explainable AI in Alzheimer's Detection}
%
\author{Yasmine Mustafa\inst{1}\orcidlink{0009-0002-3512-9659} \and
Mohamed Elmahallawy\inst{2}\orcidlink{0000-0002-5731-9253} \and
Tie Luo\inst{3}\orcidlink{0000-0003-2947-3111}\thanks{Corresponding author.}}
\authorrunning{Y. Mustafa et al.}
%
\institute{Missouri University of Science and Technology, Rolla, MO 65409, USA\\ \and
Washington State University, Richland, WA 99354, USA\\ \and
University of Kentucky, Lexington, KY 40506, USA\\
\email{yam64@mst.edu}, \email{mohamed.elmahallawy@wsu.edu}, \email{t.luo@uky.edu}}
 

\enlargethispage{3\baselineskip}

\maketitle
\renewcommand\thefootnote{}  
\footnotetext{2025 PAKDD Workshop on Pattern mining and Machine learning for Bioinformatics (PM4B).\\
\scriptsize\sf Publisher's Disclaimer: This version of the article has been accepted for publication after peer review, but is not the Version of Record and does not reflect post-acceptance improvements or any corrections (if applicable). The Version of Record is available online at: https://www.springernature.com. Use of this Accepted Version is subject to the publisher’s Accepted Manuscript terms of use https://www.springernature.com/gp/open-research/policies/accepted-manuscript-terms.}
\renewcommand\thefootnote{\arabic{footnote}}  

\begin{abstract}
Alzheimer’s disease (AD) leads to progressive cognitive decline, making early detection crucial for effective intervention. While deep learning models have shown high accuracy in AD diagnosis, their lack of interpretability limits clinical trust and adoption. This paper introduces a novel pre-model approach leveraging Jacobian Maps (JMs) within a multi-modal framework to enhance explainability and trustworthiness in AD detection. By capturing localized brain volume changes, JMs establish meaningful correlations between model predictions and well-known neuroanatomical biomarkers of AD. We validate JMs through experiments comparing a 3D CNN trained on JMs versus on traditional preprocessed data, which demonstrates superior accuracy. We also employ 3D Grad-CAM analysis to provide both visual and quantitative insights, further showcasing improved interpretability and diagnostic reliability.

\end{abstract}

\section{Introduction}

Alzheimer's disease (AD) is a leading cause of dementia, posing immense medical and economic challenges globally. Characterized by gradual cognitive decline, AD progresses from mild memory lapses to severe functional impairments. With Alzheimer's cases projected to rise significantly in the coming decades \cite{scheltens2021alzheimer}, early detection---e.g., of mild cognitive impairment (MCI)---is critical to slowing disease progression. Advances in neuroimaging and machine learning, especially with multi-modal data, have significantly improved AD diagnosis \cite{rathore2017review, mustafa2024unmasking, venugopalan2021multimodal}. However, skepticism persists due to the ``black-box'' nature of AI models, necessitating explainable AI (XAI) methods to foster trust among clinicians by elucidating the ``why'' behind predictions.

To address these challenges,  explainability in AI typically operates at three levels: pre-model, focusing on data-level insights (e.g., feature engineering to highlight key biomarkers); in-model, leveraging inherently interpretable models (e.g., decision trees or linear models); and post-model, generating explanations after predictions (e.g., heatmaps or saliency maps). 
While in-model approaches often struggle to capture complex patterns, post-model XAI tools, such as Grad-CAM \cite{selvaraju2020grad}, have gained significant attraction due to their applicability to deep learning models. However, such heatmap-based methods face key challenges in medical imaging, particularly in Alzheimer's disease: (i) the absence of ground truth for validating visual explanations against actual dementia-related regions; (ii) insufficient quantitative metrics for evaluating alignment with known pathology. Techniques like voxel-based morphometry (VBM), which identify structural brain changes through statistical methods, are often overlooked in training pipelines, leading to potentially suboptimal performance. Notably, there is a dearth of effective pre-model approaches for highlighting meaningful neuroanatomical changes before model training, which could enhance interpretability.

This paper introduces Jacobian Maps (JMs), derived from the VBM preprocessing pipeline but repurposed in this paper as a pre-model anatomical reference, to enhance explainability in AD detection. By computing Jacobian determinants across an image, JMs generate a matrix that accurately captures spatial and directional changes in medical scans, serving as a subject-specific ground truth. By pinpointing voxel-level cerebral changes relative to a healthy brain, JMs highlight morphometric variations, improving both interpretability and trust in AD detection. Our main contributions are:
\begin{itemize}
\item {\bf Enhanced interpretability with JM:} We transform brain imaging data (MRI, CT, PET) into JMs, allowing models to reveal localized volumetric changes across distinct brain regions. This improves interpretability and provides deeper insights into previously opaque model predictions.

\item {\bf Holistic methodology:} Unlike existing patch-based or slice-wise approaches that lose spatial context, our method processes full brain scans, which preserves voxel-level details. JM also eliminates the need for segmentation, reducing complexity and computational costs.

\item {\bf Improved AD detection without model modification:} As a pre-model preprocessing step, JM enhances detection accuracy across all AD stages without altering the AI model. Our approach achieves 95.2\% (CN), 96.3\% (MCI), 90.2\% (MLD), and 90.2\% (MOD), outperforming traditional methods (88.3\%, 90.5\%, 83.4\%, and 83.4\%, respectively). 

\item {\bf Comprehensive validation of interpretability:} Jacobian-derived heatmaps provide qualitative and quantitative insights into brain volumetric changes. Using the MNI152 template and Harvard-Oxford atlas, we show that the {\bf frontal-temporal region} consistently exhibits the highest intensity values across all dementia stages, underscoring its clinical relevance and our method’s ability to pinpoint key neuroanatomical patterns.

\end{itemize}

\section{Related Work}
Existing XAI methods applied to AD detection can be grouped into three main categories.

{\bf Pre-model (ante-hoc) approaches.} These methods often segment or partition the brain into smaller regions for localized analysis. For example, Amoroso et al. \cite{amoroso2023explainability} employed graph-based models, treating brain patches as network nodes connected via Pearson's correlation. This approach highlights connectivity patterns in regions like the hippocampus and amygdala for AD and the posterior cingulate for MCI to avoid computationally intensive voxel-wise analyses. Liu et al. \cite{liu2018multi} used a three-dimensional CNN to segment brain scans into patches, pre-training two networks to learn region-specific features, which were later integrated using a 2D CNN to enhance interpretability.

{\bf In-Model (intrinsic) approaches.} These techniques embed explainability into the training process. For instance, Yu et al. \cite{yu2022novel} introduced a multi-stage aggregation module to generate visual explanations during AD diagnosis. Similarly, Mustafa et al. \cite{mustafa2024unmasking} employed a Jacobian-Augmented Loss (JAL) function, utilizing Jacobian Saliency Maps (JSM) to penalize spurious predictions, ensuring model decisions are based on meaningful patterns.

{\bf Post-model (post-hoc) approaches.} These methods focus on interpreting pre-trained models. El-Sappagh et al. \cite{el2021multilayer} used a Random Forest classifier with Gini impurity and SHAP \cite{lundberg2017unified} to provide global feature importance and decision-level explanations. Kamal et al. \cite{kamal2021alzheimer} combined CNNs with gene expression data, leveraging LIME \cite{ribeiro2016should} for gene significance interpretation in AD classification.

{\bf Jacobian maps (JM):} While JM was used to analyze structural changes in AD, their potential for XAI remains untapped. Spasov et al. \cite{spasov2019parameter} used JMs to quantify local volumetric changes, while Abbas et al. \cite{abbas2023transformed} developed JD-CNN to train directly on JMs, improving spatial correlation analysis. Mustafa et al. \cite{mustafa2023diagnosing} extended this by integrating JMs from MRI and CT scans to identify regions of brain atrophy, enhancing the focus on structural changes before training. To the best of our knowledge, this paper is the first that introduces JM into XAI to explain model predictions.

\section{Method}

\subsection{Transforming Brain Images into Jacobian Maps (JM)}\label{sec:reg}

To analyze brain imaging modalities (MRI, CT, PET) and provide explainable insights, we propose transforming them into JMs to leverage the effectiveness of JM in nonlinear image registration. Image registration is a process that aligns individual scans to a standardized brain template so as to minimize individual variations and enable assessment of relative volume differences (deformations). It helps identify significant anatomical differences across populations (e.g., Alzheimer’s patients vs. healthy controls). 

Importantly, JM enables holistic brain analysis without partitioning the brain into patches or requiring segmentation like other methods, thereby minimizing data loss, complexity, and improving dementia detection accuracy. Moreover, JM facilitates feature attribution by delineating and quantifying volumetric transformations in specific brain regions, which provide a detailed pre-model explanation to guide model training.

In this work, we use the MNI152 brain template as a standard reference for image registration. Comparing an individual’s scans to this template allows detection of localized structural changes via deformation maps, while comparing such scans at different times allows assessment of disease progression. This approach, known as {\em tensor-based morphometry} \cite{riyahi2018quantifying}, is grounded in the field of computational anatomy. This framework involves transforming a source image $M$ to align with a target image $F$ using a spatial mapping $\phi$, producing a deformation vector field $\vec{v}$:
\begin{equation}
    \vec{v}(x, y, z)= \phi(x, y, z)-(x, y, z)
\end{equation}
where $\phi(x,y,z)$ represents the transformed coordinates. To ensure a smooth (differentiable), one-to-one (invertible) deformation, a regularization constraint is introduced, framed as an optimization problem minimizing a cost function:
\begin{equation}
   \ell(\phi, M \circ F) = \ell_{sim} + \alpha \times \ell_{Reg}
\end{equation}
where $\ell_{sim}$ measures how well two images align with each other, $\ell_{Reg}$ is a regularization term, and $\alpha$ balances the two terms. We use Mattes Mutual Information (MMI) \cite{mattes2003pet}  as it identifies optimal image alignment by evaluating the shared global information from the joint histogram. Additionally, MMI reduces the risk of overfitting by discouraging excessive clustering of marginal probabilities, which denotes the probability of individual variables being considered in isolation, as opposed to joint probabilities. Excessive clustering of marginal probabilities could lead to overfitting by focusing too much on specific individual features rather than the overall distribution. Unlike cross-correlation, which is affected by local intensity fluctuations and faces certain multimodality challenges, MMI excels in achieving robust rigid registration \cite{avants2008symmetric}. Thus,
\begin{equation}
\ell_{sim} = MMI(M, F) = \sum_{m\in M} \sum_{f\in F} P(m, f) \log\frac{P(m, f)}{P(m)Q(f)}
\end{equation}
where $P(m, f)$ is the joint probability of intensities $m$ in image $M$ and $f$ in image $F$. $P(m)$ and $Q(f)$ are marginal probabilities of $m$ and $f$, respectively. For $\ell_{Reg}$, we use B-spline regularization \cite{tustison2013explicit} to enforce smoothness and prevent overfitting in our model. B-splines, or basis splines, are piecewise polynomial functions that provide flexible yet stable approximations by constraining the curvature of the function, ensuring that the model generalizes well to unseen data while avoiding excessive complexity, which can be expressed as:
\begin{equation}
\ell_{Reg} = \int \left| \nabla^2 \phi(x, y, z) \right|^2 dV
\end{equation}
The Jacobian matrix $\mathcal{J}$, derived from the deformation vector field $\vec{v}(x, y, z)$, encodes local deformations such as stretching, shearing, and rotation:
\begin{equation}
\mathcal{J}(\vec{v}) =
\begin{pmatrix}
\frac{\partial \vec{v}_x}{\partial x} & \frac{\partial \vec{v}_x}{\partial y} & \frac{\partial \vec{v}_x}{\partial z} \\
\frac{\partial \vec{v}_y}{\partial x} & \frac{\partial \vec{v}_y}{\partial y} & \frac{\partial \vec{v}_y}{\partial z} \\
\frac{\partial \vec{v}_z}{\partial x} & \frac{\partial \vec{v}_z}{\partial y} & \frac{\partial \vec{v}_z}{\partial z}
\end{pmatrix}
\end{equation}
This process forms a Jacobian tensor that captures the rate of change of the vector field in three-dimensional space (width, height, and depth). The Jacobian determinant, $\text{Det}(\mathcal{J})$, computed at each voxel, quantifies the local volumetric change. By calculating these determinants across the entire image domain ($x = 1, \ldots, W; \; y = 1, \ldots, H; \; z = 1, \ldots, D$), we construct the JM, which provides a spatial representation of local deformations and volume changes as

\[
J_{\text{map}}(M) =
\begin{cases}
\text{Expansion}, & \text{if } \text{Det}(\mathcal{J}) > 1 \\
\text{No change}, & \text{if } \text{Det}(\mathcal{J}) = 1 \\
\text{Compression}, & \text{if } \text{Det}(\mathcal{J}) < 1
\end{cases}
\]

\subsection{Brain Image Processing} 

This subsection details how we handle brain imaging data, with a focus on MRI. MRI provides highly detailed visuals of progressive cerebral atrophy, especially through T1-weighted sequences. Compared to CT, MRI offers superior soft tissue contrast and multi-contrast imaging for detecting AD \cite{aramadaka2023neuroimaging}. While PET visualizes beta-amyloid plaques, MRI indirectly assesses the amyloid burden through indicators like hippocampal volume loss and provides better spatial resolution.

\textbf{Data Collection.} We utilize 1557 MRI images from the OASIS-3 dataset \cite{lamontagne2019oasis}, which comprises 1377 participants: 755 cognitively normal adults and 622 with varying cognitive decline, aged 42–95. To avoid overlap between training and test sets, images from the same subject are confined to one dataset partition, ensuring the evaluation of unseen data for accurate generalization. Clinical Dementia Rating (CDR) scores (0 to 3) guide the diagnosis, where 0 indicates no dementia, and 3 indicates severe dementia. To address the issue of limited data for severely demented cases, we merge moderate and severe stages into one category, resulting in four groups: normal (CN), mild cognitive impairment (MCI), mild dementia (MLD), and moderate-to-severe (MOD). This classification provides more granularity than binary methods (AD and non-AD).

\begin{figure*}[t]
  \centering
\centering \subfloat[\centering\scriptsize{Raw}]{\includegraphics[width=0.129\linewidth]{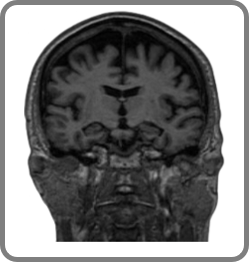}}\
\subfloat[\centering\scriptsize{Bias field correction.}]{\includegraphics[width=0.165\linewidth]{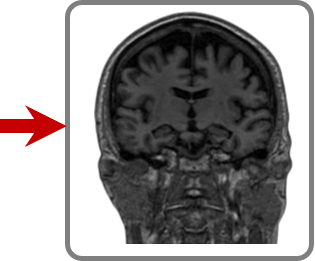}}\
   \subfloat[\centering\scriptsize{Brain extraction.}]{\includegraphics[width=0.165\linewidth]{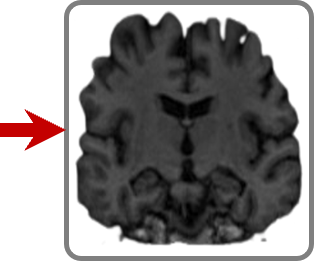}}\
\subfloat[\centering\scriptsize{Register to MNI152.}]{\includegraphics[width=0.165\linewidth]{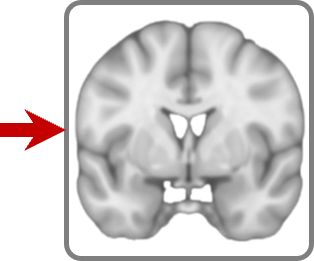}}\
   \subfloat[\centering\scriptsize{Registered.}]{\includegraphics[width=0.165\linewidth]{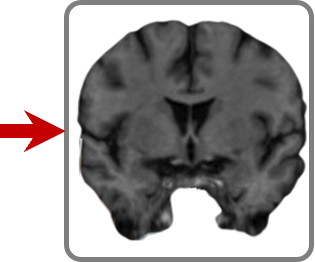}}\
     \subfloat[\centering\scriptsize{Jacobian map.}]{\includegraphics[width=0.165\linewidth]{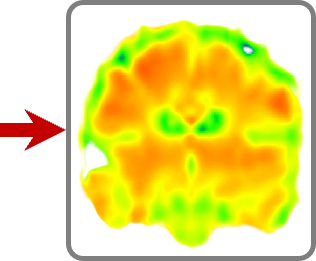}}
  \caption{Preprocessing pipeline.}
  \label{preprocessing_pipeline}
\end{figure*}

{\bf Preprocessing Pipeline.} Preprocessing is particularly critical in medical imaging. Our MRI data preprocessing pipeline is shown in \fref{preprocessing_pipeline}. Bias field correction, performed with FLIRT \cite{jenkinson2002improved}, mitigates intensity variations from magnetic field inhomogeneities, enhancing image consistency. Brain extraction using BET \cite{smith2002fast} isolates the brain region, reducing computational load and improving volumetric measurement accuracy.

The registration process employs symmetric normalization (SYN) using ANTs \cite{avants2009advanced} for global alignment through affine transformations, addressing positional and orientation differences across subjects. 
Finally, JMs, computed for each MRI, is a crucial step to identify potential indicators of neurological conditions and offer insights into brain morphology by quantifying local volume variations over distinct brain regions. Our ablation analysis (discussed in the next section) highlights the superior ability of JMs, compared to registered images without JM, to reveal patterns of brain atrophy---a hallmark of neurodegenerative disorders like AD.

{\bf Handling Imbalanced Data.} To address the class imbalance inherent in MRI-based disease classification, we use SMOTE \cite{chawla2002smote}, which generates synthetic samples by interpolating between minority class instances, enhancing dataset balance and representation.


\begin{figure}[t]
 \centering
  \includegraphics[width=1\linewidth]{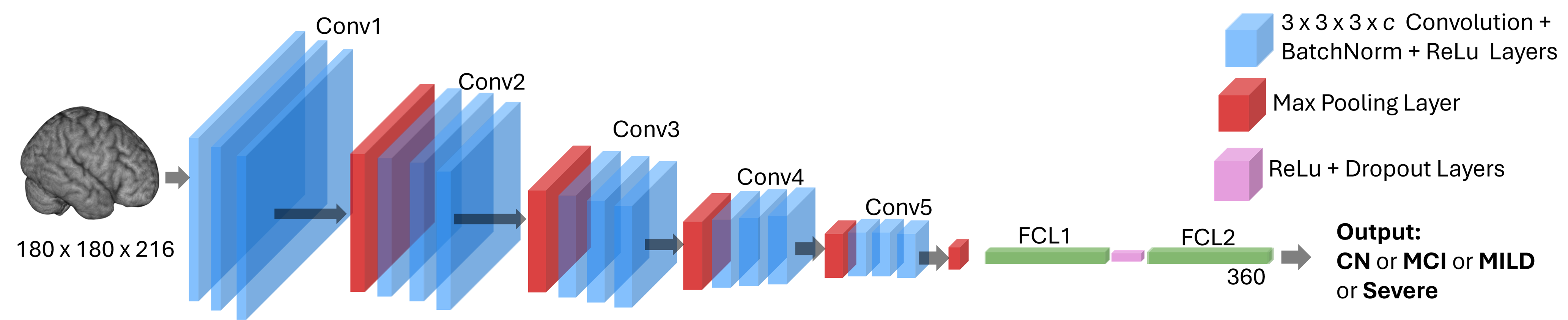}
  \caption{The 3D CNN consists of five 3D convolutional layers with kernel sizes of 3×3×3×$c$, where $c$ represents the number of input channels. The output is passed through two fully connected (FC) layers after flattening. Each conv layer is followed by batch norm and ReLU activation, max-pooling is used at selected locations, and Dropout is applied at the first FC layer for regularization. The final class logits are normalized by softmax.}
  \label{Model}
\end{figure}

\section{Experiments and Explainability Analysis}

\subsection{Model and Training}\label{sec:model}

The model used in our simulation is depicted in \fref{Model}. An ablation study by Wang et al. \cite{wang2023deep} determined the optimal model configuration for MRI analysis, identifying 44 channels per convolutional layer as yielding peak performance. To improve computational efficiency, we reduced the channels to 10; on the other hand, to handle the increased feature complexity, we increased the number of neurons in the FC layer from 64 to 360.

{\bf Training.} We employed 5-fold cross-validation, splitting the data into subsets to train for 50 epochs per fold, using early stopping to prevent overfitting. Training stops when validation loss does not decrease for a predefined patience period. Model checkpoints are saved to retain the best-performing model. The model’s hyperparameters include a batch size of 15, the Adam optimizer with a learning rate of $10^{-4}$, and the cross-entropy loss function.

\begin{figure}[t]
    \centering
     \includegraphics[width=.75\linewidth]{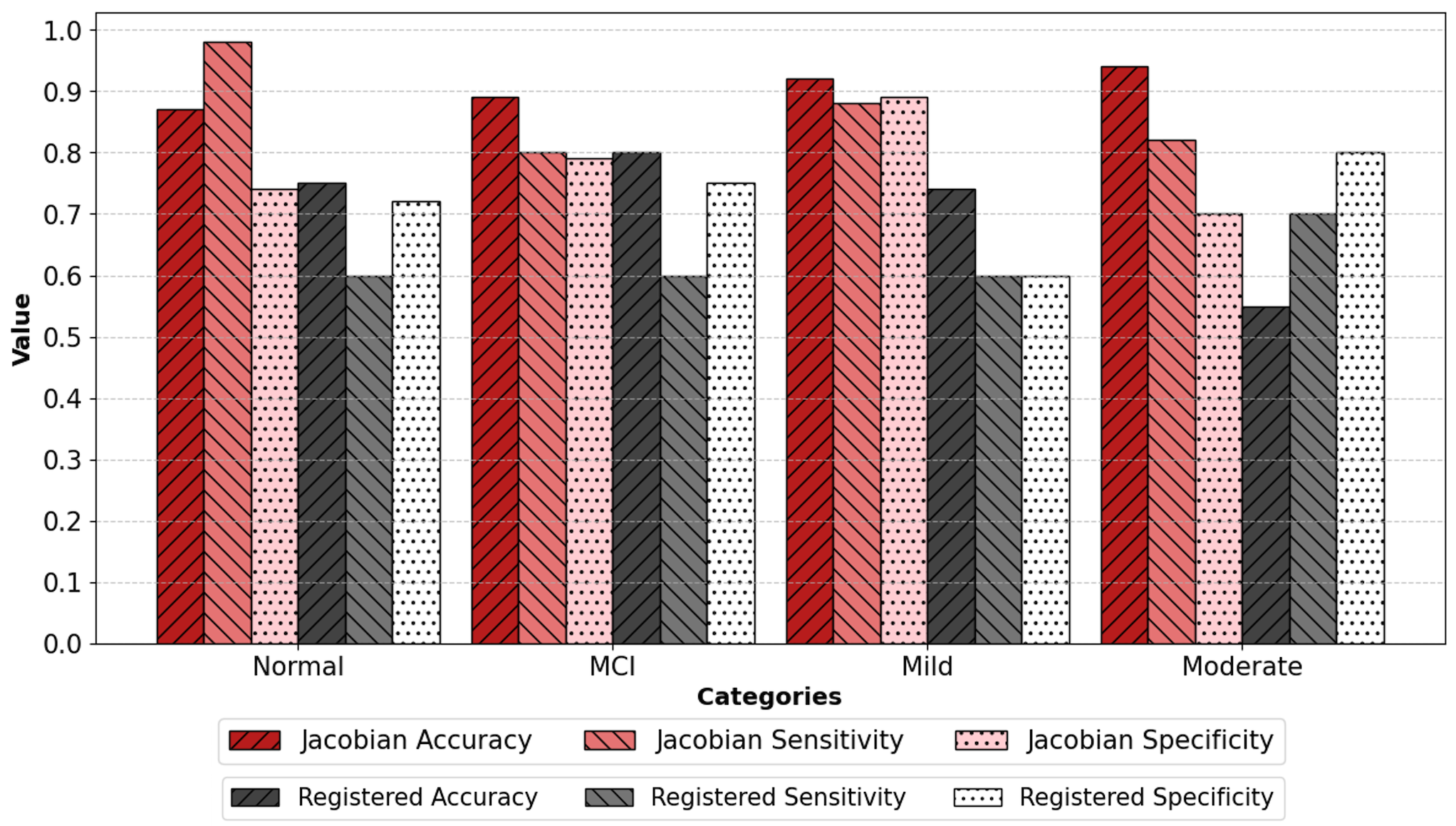}
 \caption{Performance comparison per each AD class between conventional registration (without JM; gray-scale bars) versus registration with JM (red-scale bars).}\label{bar_graph}
\end{figure}

\begin{figure*}[t]
        \centering \subfloat[\centering\scriptsize{Fold 1}]{\includegraphics[width=0.19\linewidth]{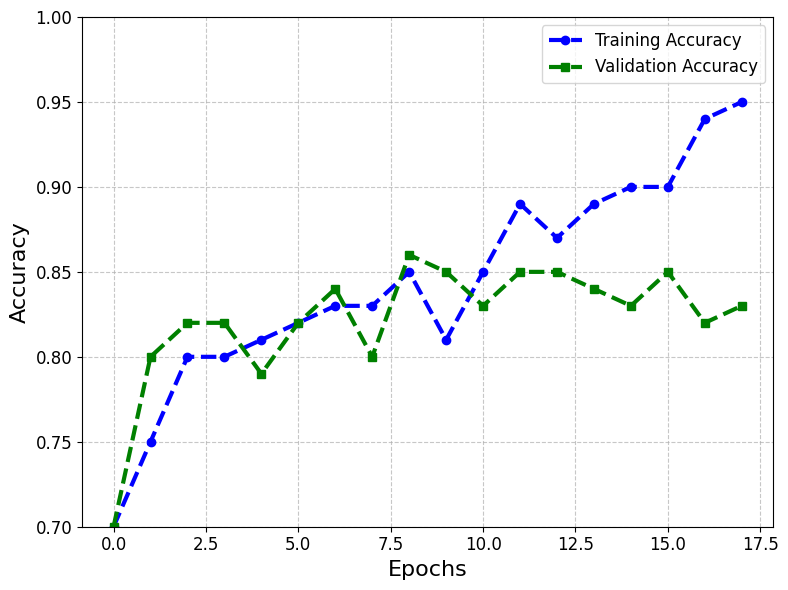}}\
        \centering \subfloat[\centering\scriptsize{Fold 2}]{\includegraphics[width=0.19\linewidth]{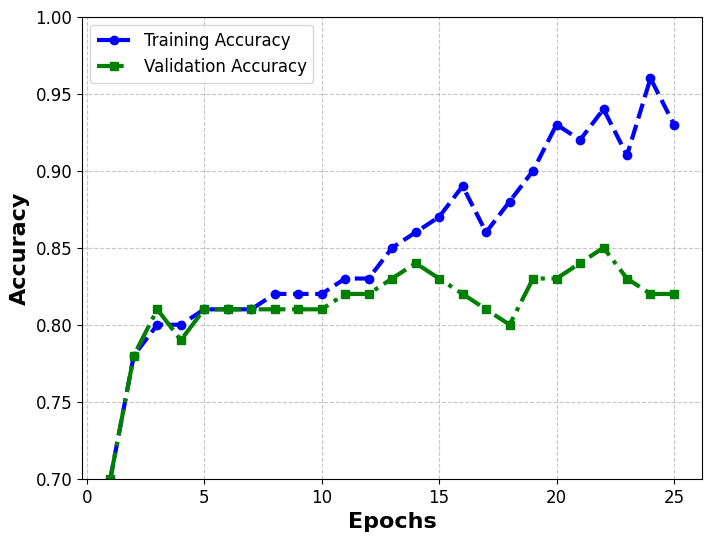}}\
          \centering \subfloat[\centering\scriptsize{Fold 3}]{\includegraphics[width=0.19\linewidth]{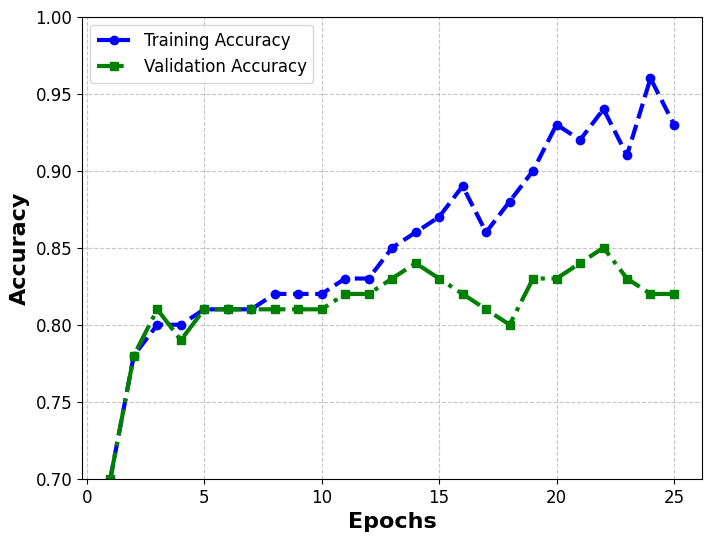}}\
            \centering \subfloat[\centering\scriptsize{Fold 4}]{\includegraphics[width=0.19\linewidth]{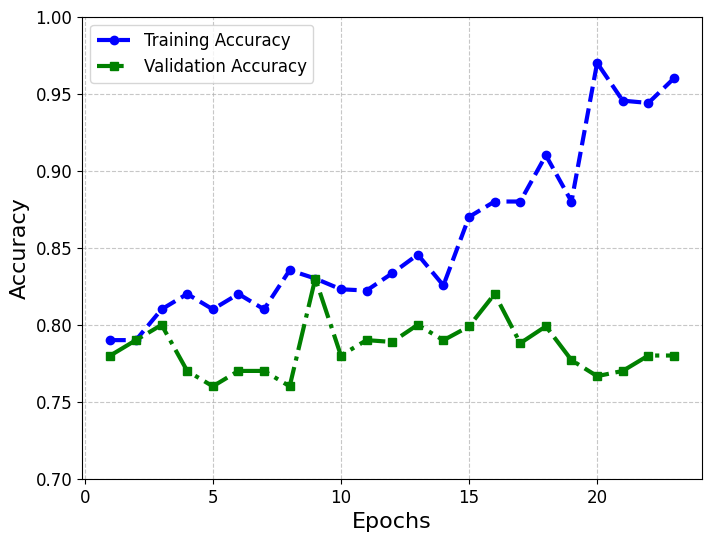}}\
              \centering \subfloat[\centering\scriptsize{Fold 5}]{\includegraphics[width=0.19\linewidth]{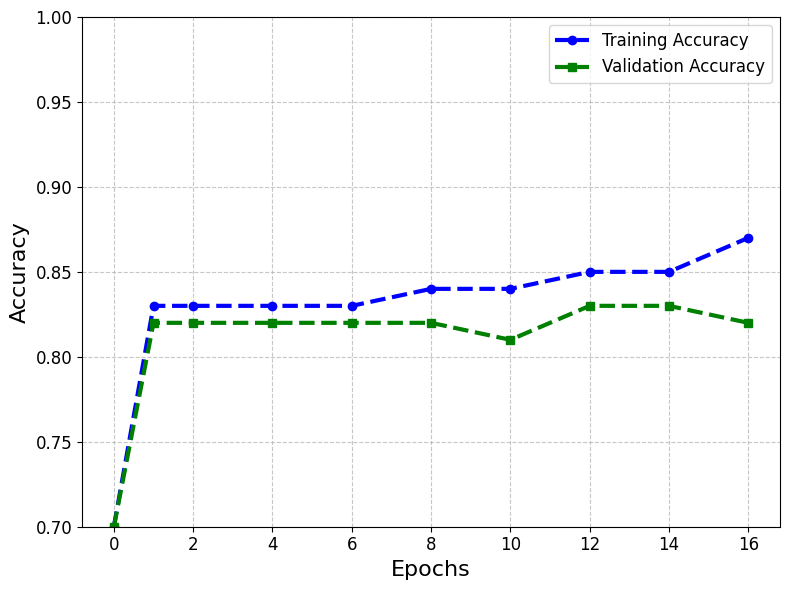}}\
              \\
        \centering \subfloat[\centering\scriptsize{Fold 1}]{\includegraphics[width=0.19\linewidth]{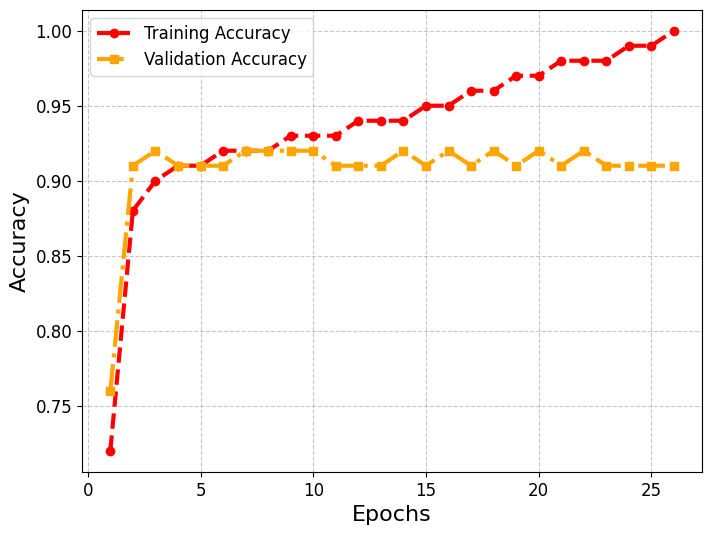}}\
        \centering \subfloat[\centering\scriptsize{Fold 2}]{\includegraphics[width=0.19\linewidth]{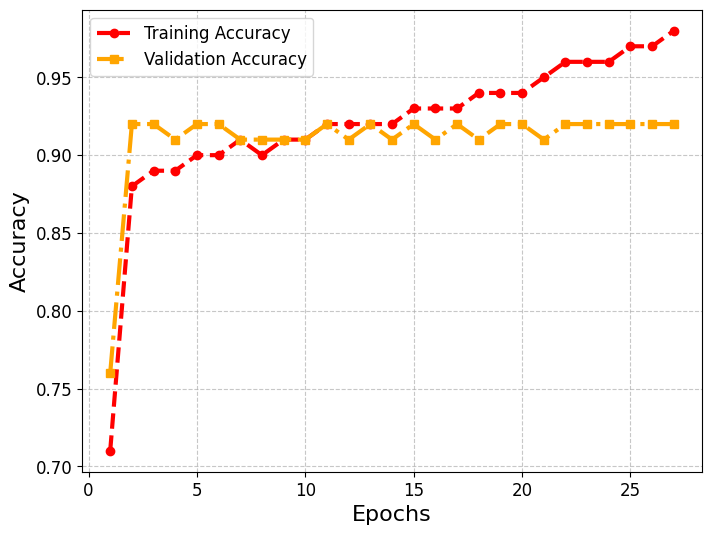}}\
          \centering \subfloat[\centering\scriptsize{Fold 3}]{\includegraphics[width=0.19\linewidth]{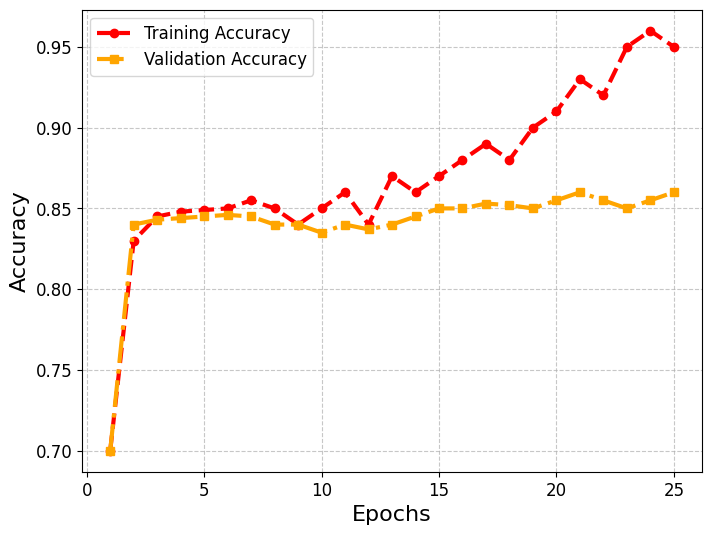}}\
            \centering \subfloat[\centering\scriptsize{Fold 4}]{\includegraphics[width=0.19\linewidth]{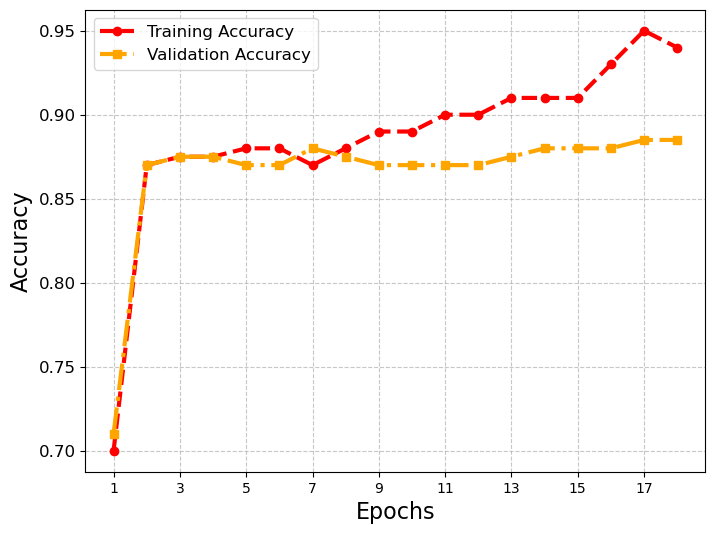}}\
              \centering \subfloat[\centering\scriptsize{Fold 5}]{\includegraphics[width=0.19\linewidth]{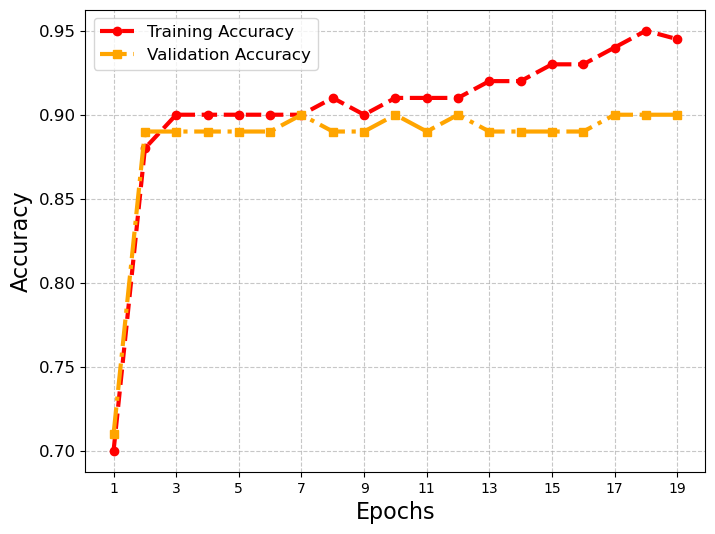}}\
 \caption{Learning curves of training and validation accuracy across 5-fold cross-validation. Upper row: {\bf without JM}, Lower row: {\bf with JM}, showing better stability and convergence.}

    \label{performance_curves_1}
\end{figure*}

\subsection{Model Performance Evaluation}

\fref{bar_graph} compares the model's performance using metrics such as accuracy, precision, and recall across registered images and JMs. The cross-validation, applied across AD classes (Normal, MCI, Mild, and Moderate), shows that JMs consistently outperform registered images in all metrics. These results underscore the effectiveness of JMs in capturing informative local volumetric changes (expansion/compression) in brain anatomy, which enhances model's discriminative power and leads to improved model performance and diagnostic accuracy.

\fref{performance_curves_1} further illustrate the learning curves for models trained on registered images versus JMs. Models trained on JM exhibit smoother learning curves and stabler validation accuracy compared to those trained on conventionally registered images. The top row (no JM) shows greater fluctuations, while the bottom row (with JM) demonstrates more consistent convergence in addition to improved training and validation performance, highlighting the benefits of JM's enhanced feature representation.

\subsection{Interoperability Evaluation}

\subsubsection{\bf 1) Qualitative Analysis.} 

\begin{wrapfigure}{r}{0.675\textwidth} 
\begin{minipage}{0.675\textwidth}
\begin{algorithm}[H]
\caption{Compute 3D Grad-CAM}
   \begin{algorithmic}[1]
\State \textbf{Step 1:} Calculate the gradient of the score $y^C$ for class $C$ with respect to the feature map activation of the unit in the final convolutional layer at location $(i, j, m)$ in three dimensions: $\frac{\partial y^C}{\partial A_u}$
\State \textbf{Step 2:} Obtain the importance weights of the unit by Global Average Pooling (GAP) of the gradients:
\Statex \hspace{2em} $\lambda_u^C = \frac{1}{Z} \sum_{i,j,m} \frac{\partial y^C}{\partial A_u}$,
\Statex where $Z = u \times v \times w$, and $A_u(i,j,m)$ represents the activation at location $(i, j, m)$.
\State \textbf{Step 3:} Compute a weighted sum across all units and apply the Rectified Linear Unit (ReLU) activation to the result to eliminate features that negatively impact the class prediction:
\Statex \hspace{2em} $\operatorname{ReLU}\left(\sum_u \lambda_u^C A_u\right)$
\State \textbf{Step 4:} Upsample the resulting map to the original input size.
   \end{algorithmic}
\label{3D_CAM_algorithm}
\end{algorithm}
\end{minipage} 
\end{wrapfigure}

To harness the interpretive capability of JMs, we employ Gradient-weighted Class Activation Mapping (Grad-CAM) \cite{selvaraju2020grad}, which visualizes and interprets the contributions of specific regions in 3D CNN predictions. Grad-CAM computes the gradients of the target class logits with respect to feature maps from the last convolutional layer. These gradients are aggregated to assign weights to each feature map, producing a Class Activation Map (CAM) (see Algorithm \ref{3D_CAM_algorithm}). The CAM is superimposed on the original image, visually highlighting the regions most relevant to the model’s prediction.

\begin{figure}[t]
\begin{subfigure}[b]{0.49\textwidth}
         \centering
      \includegraphics[width=1\linewidth]{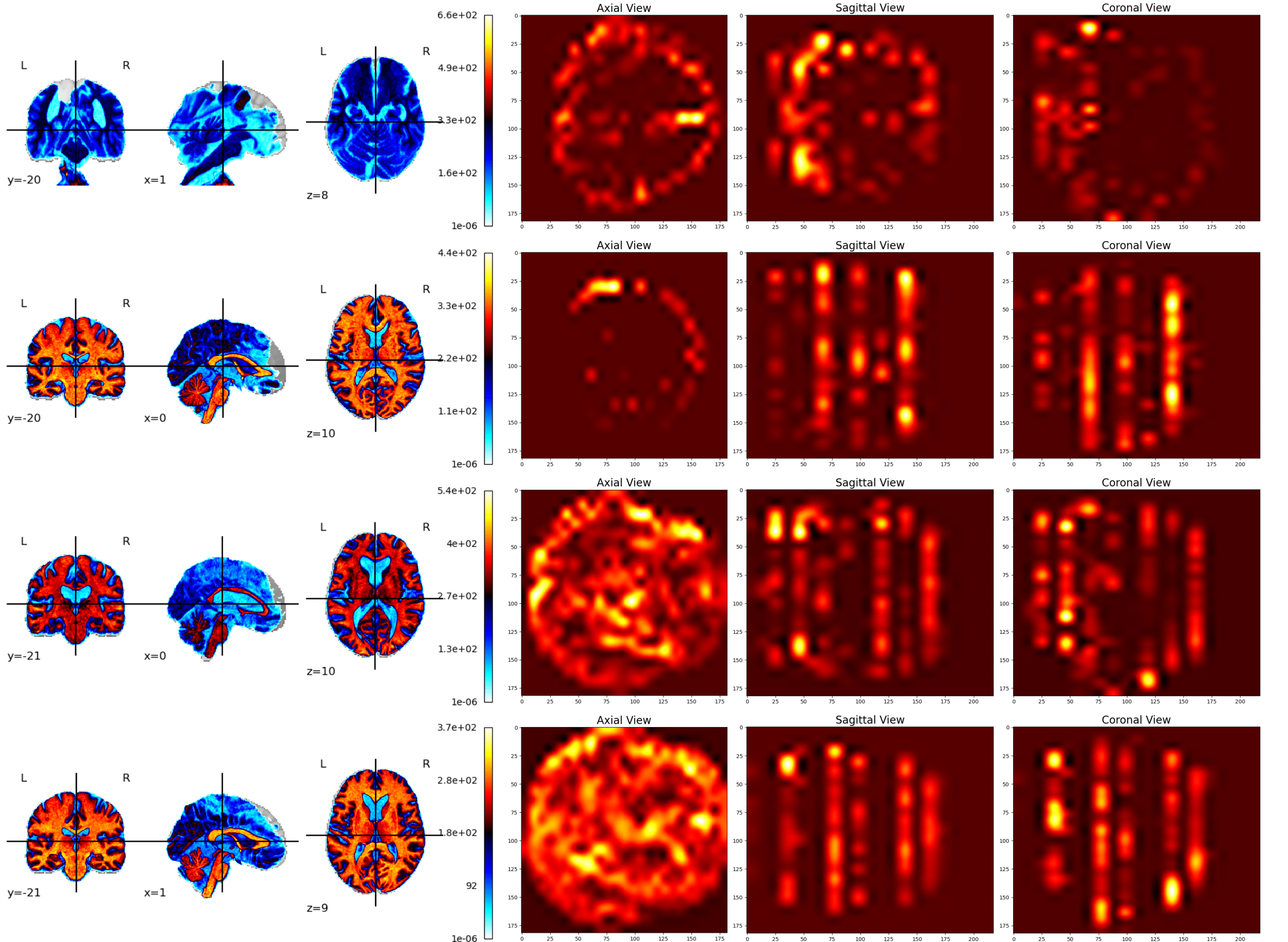}
    \caption{\centering }
     \end{subfigure}
     \hfill
\begin{subfigure}[b]{0.49\textwidth}
         \centering
      \includegraphics[width=1\linewidth]{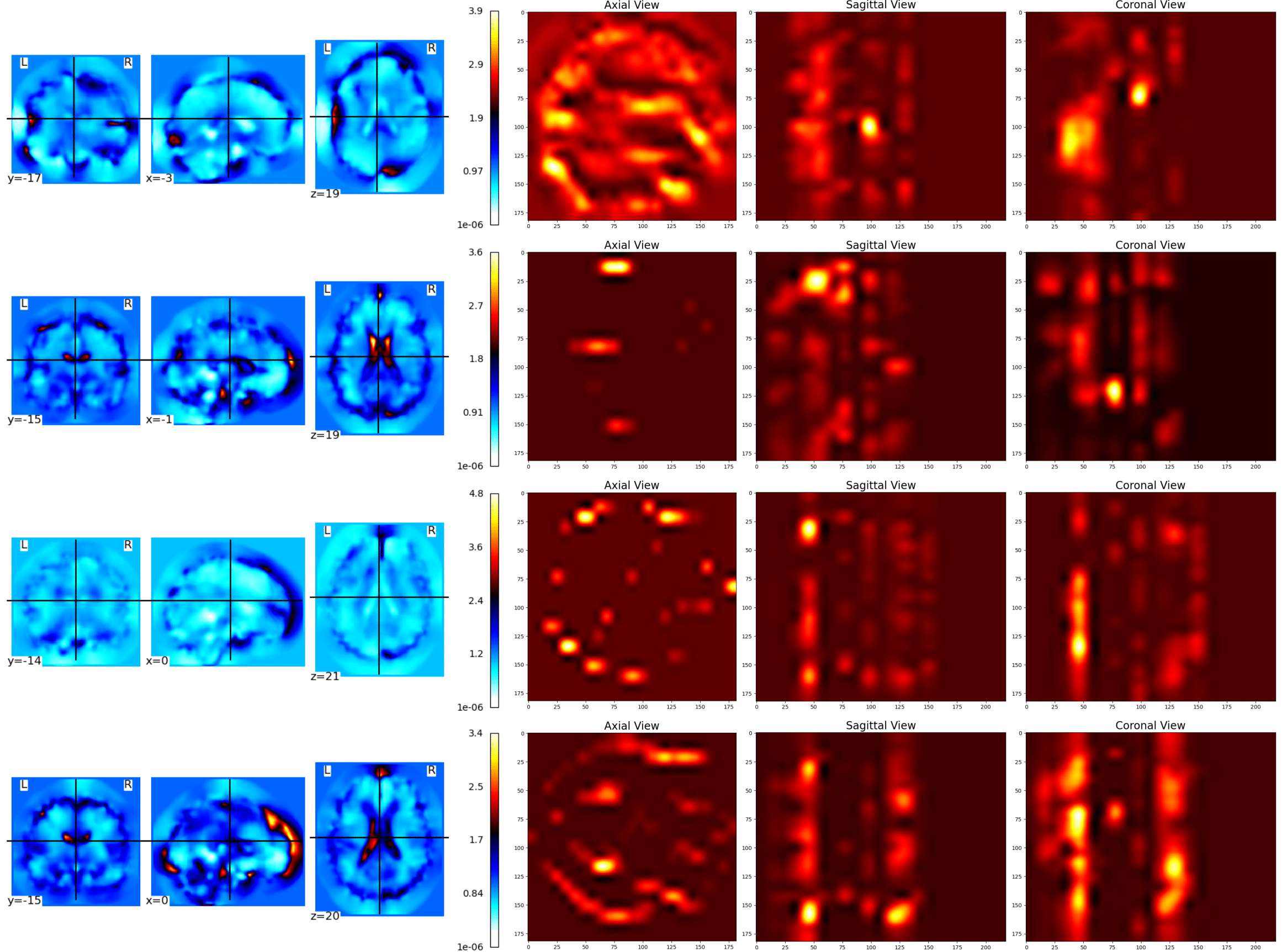}
    \caption{\centering }
     \end{subfigure}
 \caption{Heatmaps (2\textsuperscript{nd} and 4\textsuperscript{th} groups of columns) generated by 3D Grad-CAM after the 3\textsuperscript{rd} conv block of our 3D CNN for four brain MRIs, registered (a) without JMs and (b) with JMs. JMs enhance interpretability by providing more specific and localized visualizations of structural brain changes, enabling precise identification of subtle yet critical variations.}
   \label{heatmaps}
\end{figure}

When applied to JMs, the highlighted regions correspond directly to areas of volumetric compression and expansion in the brain. As shown in \fref{heatmaps}, JM-derived heatmaps excel at pinpointing specific regions of interest compared to those generated from conventional registered images. While heatmaps from registered images provide broader, less targeted representations, they often fail to capture the specificity required to identify significant volumetric changes. In contrast, JM-derived heatmaps deliver focused and detailed visualizations, revealing localized structural alterations in the brain. This enhanced specificity enables the precise identification of subtle yet critical variations, improving both model interpretability and diagnostic utility.

\begin{table}[t]
    \centering
    \caption{Brain regions ranked by importance for each AD class based on heatmap intensity.}\label{brain_regions}
    \begin{tabular}{p{3.5cm}|p{3.5cm}|p{3.5cm}|p{3.5cm}}
        \toprule
       \textbf{CN} & \textbf{MCI} & \textbf{MLD} & \textbf{MOD} \\
        \midrule
       Frontal-Temporal (2.71) & Frontal-Temporal (2.45) & Frontal-Temporal (2.76) & Frontal-Temporal (2.76) \\
         Sub-lobar (2.51) &  Temporal Lobe (1.94) &  Temporal Lobe (2.33) &  Frontal Lobe (2.28) \\
         Temporal Lobe (2.43) &  Frontal Lobe (1.89) &  Frontal Lobe (2.28) &  Parietal Lobe (1.77) \\
         Limbic Lobe (2.40) &  Sub-lobar (1.78) &  Sub-lobar (2.20) &  Limbic Lobe (2.01) \\
         Frontal Lobe (2.37) &  Background (1.73) &  Occipital Lobe (2.02) &  Occipital Lobe (2.02) \\
         Midbrain (2.29) &  Limbic Lobe (1.67) &  Pons (1.82) &  Pons (1.82) \\
         Pons (2.26) &  Occipital Lobe (1.59) &  Posterior Lobe (1.91) &  Posterior Lobe (1.91) \\
         Background (2.08) &  Anterior Lobe (1.53) &  Background (1.73) &  Background (1.73) \\
         Parietal Lobe (2.07) &  Medulla (1.37) &  Anterior Lobe (1.53) &  Medulla (1.43) \\
         Posterior Lobe (1.98) &  Midbrain (1.61) &  Medulla (1.43) &  Anterior Lobe (1.53) \\
         Medulla (1.76) &  Frontal-Temporal (2.45) &  Parietal Lobe (1.77) &  Midbrain (1.90) \\
         Anterior Lobe (1.97) &  Parietal Lobe (1.33) &  Frontal Lobe (1.89) &  Pons (1.82) \\
          \bottomrule
    \end{tabular}
\end{table}

In summary, JMs play a pivotal role in unraveling the decision-making processes of models by identifying distinct regions within an input image and quantifying their volumetric transformations. This enables precise feature attribution, assigning significance and influence to specific input regions that impact the model’s output. Consequently, JMs offer a comprehensive {\em pre-model (ante-hoc)} explanation, enhancing our understanding of how input data is interpreted by the model.

\subsubsection{\bf 2) Quantitative Analysis.} 

To quantitatively assess the model's interpretability, we align Grad-CAM-generated heatmaps with the MNI152 template by computing registration transformations. Using the template as a fixed image and each subject's MRI as a moving image, these transformations ensure consistent spatial coordinates across subjects and images. The Harvard-Oxford cortical atlas \cite{de2023explainable} is then used to calculate the average voxel intensity within each brain region for each heatmap, providing a region-based metric of activation. Regions are ranked in descending order of importance across subjects for each heatmap class. This approach identifies the brain regions most sensitive to input variations for each AD class, as shown in Table \ref{brain_regions} for JM-derived heatmaps.

The rankings in Table \ref{brain_regions} highlight which brain regions contribute most significantly to distinguishing between brain states (Normal, MCI, Mild-Moderate, and Severe). Several key observations can be made:
\begin{itemize}
    \item  {\bf Normal  State:}
 
 \begin{itemize}
     \item The {\bf Frontal-Temporal} region (2.71) is the most significant, followed by the {\bf Sub-lobar} region (2.51) and {\bf Temporal Lobe} (2.43).
     \item The {\bf Frontal-temporal} region is involved in memory, decision-making and language, and is known to degenerate early in AD. This explains its highest rank across {\bf all stages}.
 \end{itemize}

\item { \bf MCI  State:}
\begin{itemize}
    \item The {\bf Frontal-Temporal} region remains the most significant (2.45), but the {\bf Temporal Lobe} (1.94) surpasses the {\bf Sub-lobar} region (1.78) in importance.
\end{itemize}

\item {\bf Mild State:}
\begin{itemize}
    \item The {\bf Frontal-Temporal} region remains dominant (2.76), while the {\bf Temporal Lobe} (2.33) becomes more important than the {\bf Sub-lobar} region (2.20).
    \item The increasing prominence of the {\bf Temporal lobe} during the MCI and MLD stages is because this lobe includes hippocampus and entorhinal cortex, which are among the first regions to show atrophy, with memory loss being a key symptom.
\end{itemize}

\item {\bf Moderate State:}
\begin{itemize}
    \item The {\bf Frontal-Temporal} region retains its significance (2.76), but the {\bf Frontal Lobe} (2.28) and {\bf Parietal Lobe} (1.77) emerge as more critical, surpassing the {\bf Temporal Lobe} (2.02) and {\bf Limbic Lobe} (2.01).
    \item The {\bf Frontal} and {\bf Parietal Lobes} are responsible for planning, judgment, spatial orientation, and attention. Our finding shows that these areas are less impacted in early stages yet become more prominently affected in later stages as neurodegeneration spreads.
\end{itemize}
\end{itemize}

These findings highlight the dynamic changes in brain region importance as Alzheimer's disease progresses from Normal to MCI, Mild, and Moderate states. Importantly, they are {\bf consistent with clinical evidence}, which underscores the significance and validity of our XAI method.

\begin{figure}[t]
  \centering
  \subfloat[\centering\scriptsize{Jacobian Maps (axial view).}]{\includegraphics[width=0.8\linewidth]{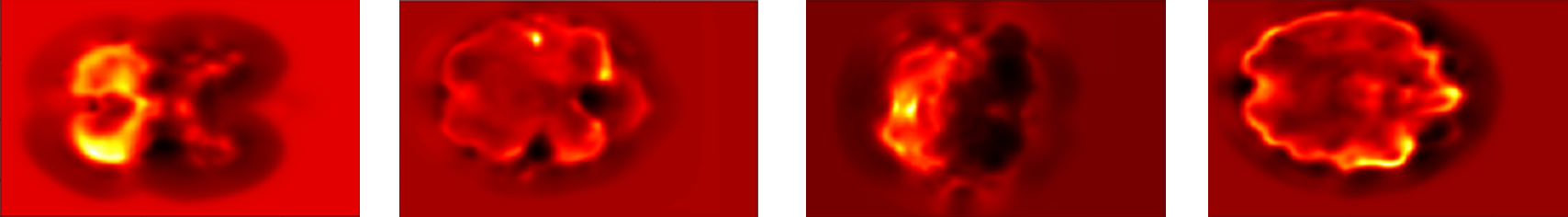}}\\
  \subfloat[\centering\scriptsize{Heatmaps (axial view).}]{\includegraphics[width=0.8\linewidth]{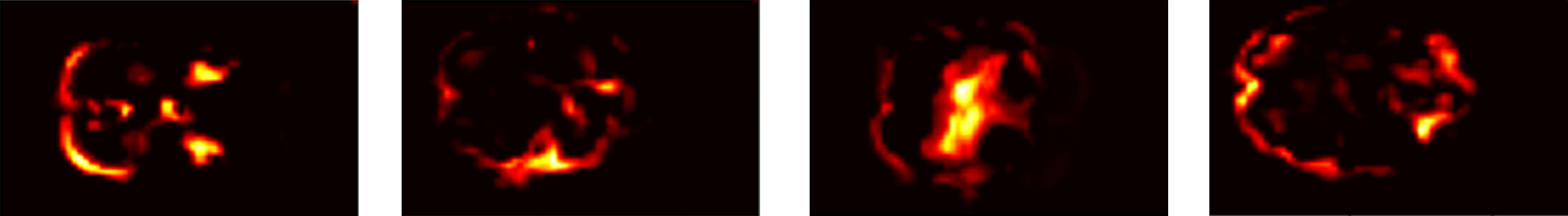}}\\
\subfloat[\centering\scriptsize{Jacobian Maps (coronal view).}]{\includegraphics[width=0.8\linewidth]{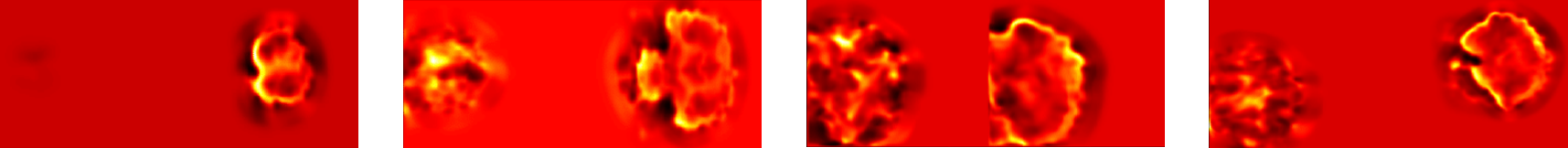}}\\
\subfloat[\centering\scriptsize{Heatmaps (coronal view).}]{\includegraphics[width=0.8\linewidth]{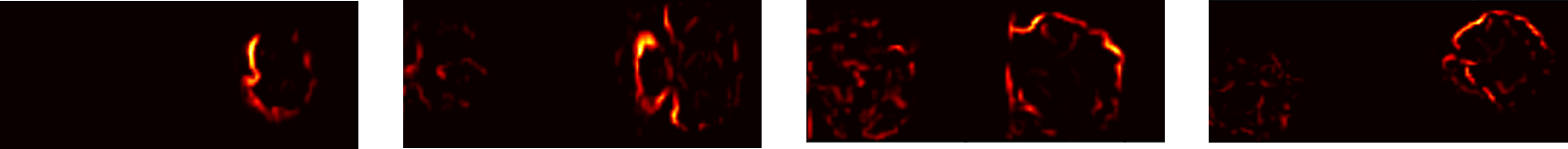}}\
   \subfloat[\centering\scriptsize{Jacobian Maps (sagital view).}]{\includegraphics[width=0.8\linewidth]{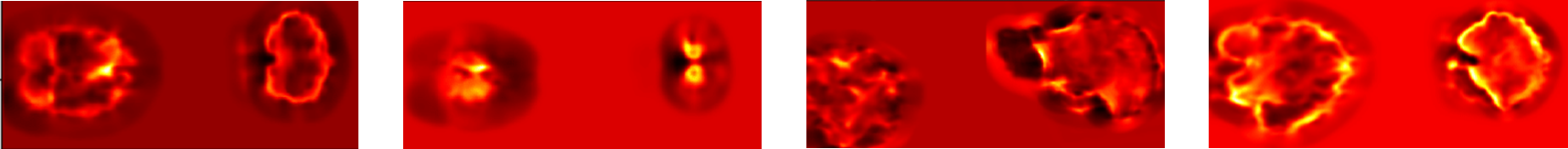}}\
\subfloat[\centering\scriptsize{Heatmaps (sagital view).}]{\includegraphics[width=0.8\linewidth]{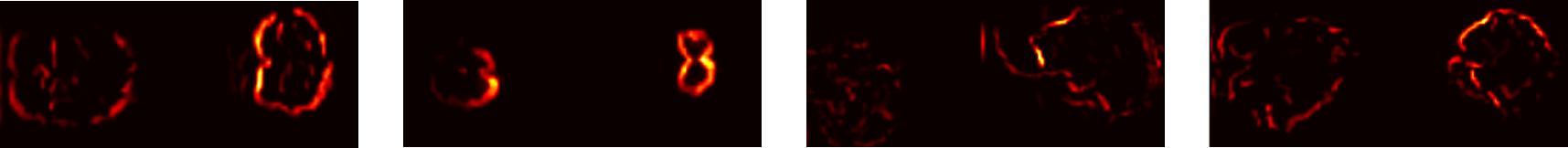}}\
  \caption{Heatmaps for fused MRI-CT brain images. }
  \label{ctmri}
\end{figure}

\subsection{Extension to Multi-modal Setting}

We extend our research into a multimodal setting by integrating MRI and CT data from the OASIS-3 dataset using an early fusion approach. Early fusion combines information from different imaging modalities at the initial stage of processing, allowing the model to leverage complementary information for improved performance. In our method, this integration involves concatenating MRI and CT data before inputting them into our model.

We employ the same model architecture outlined in Section \ref{sec:model}, with an adaptation of two additional convolutional layers to optimize computational efficiency in handling the increased dimensionality from multimodal fusion. This adjustment allows the model to more effectively capture the richer feature representations offered by both MRI and CT data.

Figure \ref{ctmri} illustrates the heatmaps generated from the combined MRI and CT features, providing insights into the spatial contributions of both modalities. Table \ref{tablemrict} presents the detection results of the 3D CNN, highlighting significant improvements when incorporating JMs compared to using registered images alone. These results confirm the efficacy of JM in enhancing interpretability and diagnostic accuracy, even in multimodal settings.

\begin{table*}[!t]
\centering
\begin{threeparttable}
\caption{Multimodal: Ablation for REG (without JM) vs. JM}\label{tablemrict}
\begin{tabular}{lcccccccccccc}
\toprule
\multirow{2}{*}{\textbf{}} & \multicolumn{4}{c}{\textbf{Accuracy}} & \multicolumn{4}{c}{\textbf{Precision}} & \multicolumn{4}{c}{\textbf{Recall}} \\
\cmidrule(lr){2-5} \cmidrule(lr){6-9} \cmidrule(lr){10-13}
\scriptsize & \scriptsize \textbf{CN} & \scriptsize \textbf{MCI} & \scriptsize \textbf{MLD} & \scriptsize \textbf{MOD} 
       & \scriptsize \textbf{CN} & \scriptsize \textbf{MCI} & \scriptsize \textbf{MLD} & \scriptsize \textbf{MOD} 
       & \scriptsize \textbf{CN} & \scriptsize \textbf{MCI} & \scriptsize \textbf{MLD} & \scriptsize \textbf{MOD} \\
\midrule
\textbf{REG} & 88.3 & 90.5 & 83.4 & 83.4 & 86.8 & 82.8 & 80 & 95.5 & 83.3 & 64.0 & 69.3 & 84.4 \\
\textbf{JM} & 95.2 & 96.3 & 90.2 & 90.2 & 92.8 & 100 &  83.33 & 98.6 & 94.96 & 89.6 & 78.6 & 90.2 \\
\bottomrule
\end{tabular}
\end{threeparttable}
\begin{tablenotes}\centering 
\item[*] {\scriptsize CN - Cognitively Normal, MCI - Mild Cognitive Impairment, MOD - Moderate-to-severe Dementia, REG - Registered without JM.}
\end{tablenotes}
\end{table*}

\section{Conclusion}

Alzheimer’s Disease (AD) remains a major challenge in neurodegenerative research, with its progressive nature causing significant cognitive, functional, and behavioral decline. Early and accurate detection is crucial for timely intervention, yet the clinical adoption of deep learning models has been hindered by their lack of interpretability and transparency.

In this paper, we introduced a novel pre-modeling approach leveraging Jacobian Maps (JMs) within a multimodal framework to enhance explainability in AD detection. By capturing localized volumetric changes across distinct brain regions, JMs provide interpretable and intuitive insights into structural brain alterations associated with AD.

Integrated into a 3D CNN, JM-based features demonstrated superior diagnostic performance across all AD stages, outperforming traditional preprocessing methods. Furthermore, combining 3D Grad-CAM with JMs enabled both visual and quantitative explanations, bridging the gap between model predictions and clinical understanding. These explanations effectively highlighted key brain regions, particularly the frontal-temporal areas, reinforcing the clinical relevance of our approach.

Moving forward, we aim to extend this methodology to other neurodegenerative disorders and explore its practical application in real-world clinical settings to further enhance trust and adoption in medical diagnostics.

%
%
%
\bibliographystyle{unsrt}
\bibliography{mybibliography}

\end{document}